\documentclass[10pt,twocolumn,letterpaper]{article}

\usepackage{cvpr}
\usepackage{times}
\usepackage{epsfig}
\usepackage{graphicx}
\usepackage{amsmath}
\usepackage{amssymb}

\usepackage{multirow}
\usepackage{booktabs}

\usepackage{amsmath}
\usepackage{mathtools}
\usepackage{algorithm}
\usepackage[noend]{algpseudocode}

\newlength{\arrow}
\newcommand*{\myrightarrow}[1]{\xrightarrow{\mathmakebox[\arrow]{#1}}}

\usepackage[pagebackref=true,breaklinks=true,letterpaper=true,colorlinks,bookmarks=false]{hyperref}

\cvprfinalcopy 

\def\cvprPaperID{4595} 

\ifcvprfinal\fi
\begin{document}

\title{Learning to Quantize Deep Networks by\\ Optimizing Quantization Intervals with Task Loss}

\author{
Sangil Jung\textsuperscript{1}\thanks{These two authors contributed equally.}\qquad
Changyong Son\textsuperscript{1}\footnotemark[1]\qquad
Seohyung Lee\textsuperscript{1}\qquad
Jinwoo Son\textsuperscript{1}\qquad
Jae-Joon Han\textsuperscript{1}\\
Youngjun Kwak\textsuperscript{1}\qquad
Sung Ju Hwang\textsuperscript{2}\qquad
Changkyu Choi\textsuperscript{1}\\\\
\textsuperscript{1}Samsung Advanced Institute of Technology (SAIT), South Korea\\
\textsuperscript{2}Korea Advanced Institute of Science and Technology (KAIST), South Korea\\
}

\maketitle

\begin{abstract}
Reducing bit-widths of activations and weights of deep networks makes it efficient to compute and store them in memory, which is crucial in their deployments to resource-limited devices, such as mobile phones. However, decreasing bit-widths with quantization generally yields drastically degraded accuracy. To tackle this problem, we propose to \textbf{learn to quantize} activations and weights via a trainable quantizer that transforms and discretizes them. Specifically, we parameterize the quantization intervals and obtain their optimal values by directly minimizing the task loss of the network. This quantization-interval-learning (QIL) allows the quantized networks to maintain the accuracy of the full-precision (32-bit) networks with bit-width as low as 4-bit and minimize the accuracy degeneration with further bit-width reduction (i.e., 3 and 2-bit). Moreover, our quantizer can be trained on a heterogeneous dataset, and thus can be used to quantize pretrained networks without access to their training data. We demonstrate the effectiveness of our trainable quantizer on ImageNet dataset with various network architectures such as ResNet-18, -34 and AlexNet, on which it outperforms existing methods to achieve the state-of-the-art accuracy.

\end{abstract}


\section{Introduction}

Increasing the depth and width of a convolutional neural network generally improves its accuracy \cite{simonyan2014very, he2016deep} in exchange for the increased memory and the computational cost. Such a memory- and computation- heavy network is difficult to be deployed to resource-limited devices such as mobile phones. Thus, many prior work have sought various means to reduce the model size and computational cost, including the use of separable filters \cite{iandola2016squeezenet, howard2017mobilenets, peng2017large}, weight pruning~\cite{han2015deep} and bit-width reduction of weights or activations \cite{zhou2016dorefa, rastegari2016xnor, choi2018pact, zhang2018lq, wang2018two, zhuang2018towards}. Our work aims to reduce bit-widths of deep networks both for weights and activations, while preserving the accuracy of the full-precision networks.

Reducing bit-width inherently includes a quantization process which maps continuous real values to discrete integers. Decrease in bit-width of deep networks naturally increases the quantization error, which in turn causes accuracy degeneration, and to preserve the accuracy of a full-precision network, we need to reduce the quantization error.  For example, Cai \etal \cite{cai2017deep} optimize the activation quantizer by minimizing mean-squared quantization error using Lloyd's algorithm with the assumption of half-wave Gaussian distribution of the response map, and some other work approximate the layerwise convolutional outputs \cite{rastegari2016xnor, li2016ternary}. However, while such quantization approaches may accurately approximate the original distribution of the weights and activations, there is no guarantee that they will be beneficial toward suppressing the prediction error from increasing. To overcome this limitation, our trainable quantizer approximates neither the weight/activation values nor layerwise convolutional outputs. Instead, we quantize the weights and activations for each layer by directly minimizing the \emph{task loss} of the networks, which helps it to preserve the accuracy of the full-precision counterpart.

Our quantizer can be viewed as a composition of a \emph{transformer} and a \emph{discretizer}. The transformer is a (non-linear) function from the unbounded real values to the normalized real values (i.e., $\mathbb{R}_{(-\infty,\infty)} \rightarrow \mathbb{R}_{[-1,1]}$), and the discretizer maps the normalized values to integers (i.e., $\mathbb{R}_{[-1,1]} \rightarrow \mathbb{I}$). Here, we propose to parameterize quantization intervals for the transformer, which allows our quantizer to focus on appropriate interval for quantization by pruning (less important) small values and clipping (rarely appeared) large values.

Since reducing bit-widths decreases the number of discrete values, this will generally cause the quantization error to increase. To maintain or increase the quantization resolution while reducing the bit-width, the interval for quantization should be compact. On the other hand, if the quantization interval is too compact, it may remove valid values outside the interval which can degrade the performance of the network. Thus, the quantization interval should be selected as compact as possible according to the given bit-width, such that the values influencing the network accuracy are within the interval. Our trainable quantizer adaptively finds the optimal intervals for quantization that minimize the task loss. 

Note that our quantizer is applied to both weights and activations for each layer. Thus, the convolution operation can be computed efficiently by utilizing bit-wise operations which is composed of logical operations (i.e., 'AND' or 'XNOR') and bitcount \cite{zhou2016dorefa} if the bit-widths of weights and activations become low enough. The weight and activation quantizers are jointly trained with full-precision model weights. Note that the weight quantizers and full-precision model weights are kept and updated only in the training stage; at the inference time, we drop them and use only the quantized weights.

We demonstrate our method on the ImageNet classification dataset \cite{russakovsky2015imagenet} with various network architectures such as ResNet-18, -34 and AlexNet. Compared to the existing methods on weight and activation quantization, our method achieves significantly higher accuracy, achieving the state-of-the-art results. Our quantizers are trained in end-to-end fashion without any layerwise optimization \cite{wang2018two}. 

In summary, our contributions are threefold:
\begin{itemize}
\item We propose a trainable quantizer with \emph{parameterized intervals} for quantization, which simultaneously performs both \emph{pruning} and \emph{clipping}.
\item We apply our trainable quantizer to both the weights and activations of a deep network, and optimize it along with the target network weights for the \emph{task-specific loss} in an end-to-end manner.
\item We experimentally show that our quantizer achieves the state-of-the-art classification accuracies on ImageNet with extremely \emph{low bit-width} (2, 3, and 4-bit) networks, and achieves high performance even when trained with a \emph{heterogeneous} dataset and applied to a pretrained network. 
\end{itemize}

\section{Related Work}
\label{sec2}
Our quantization method aims to obtain low-precision networks. Low-precision networks have two benefits: model compression and operation acceleration. Some work compresses the network by reducing bit-width of model weights, such as BinaryConnect (BC) \cite{courbariaux2015binaryconnect}, Binary-Weight-Network (BWN) \cite{rastegari2016xnor}, Ternary-Weight-Network (TWN) \cite{li2016ternary} and Trained-Ternary-Quantization (TTQ) \cite{zhu2016trained}. BC uses either deterministic or stochastic methods for binarizing the weights which are $\{-1,+1\}$. BWN approximates the full-precision weights as the scaled bipolar($\{-1,+1\}$) weights, and finds the scale in a closed form solution. TWN utilizes the ternary($\{-1,0,+1\}$) weights with scaling factor which is trained in training phase. TTQ uses the same quantization method but the different scaling factors are learned on the positive and negative sides. These methods solely considers quantization of network weights, mostly for binary or ternary cases only, and do not consider quantization of the activations. 

\begin{figure*}[h!t]
  \centering
\begin{tabular}{cc}
\includegraphics[width=0.3\linewidth]{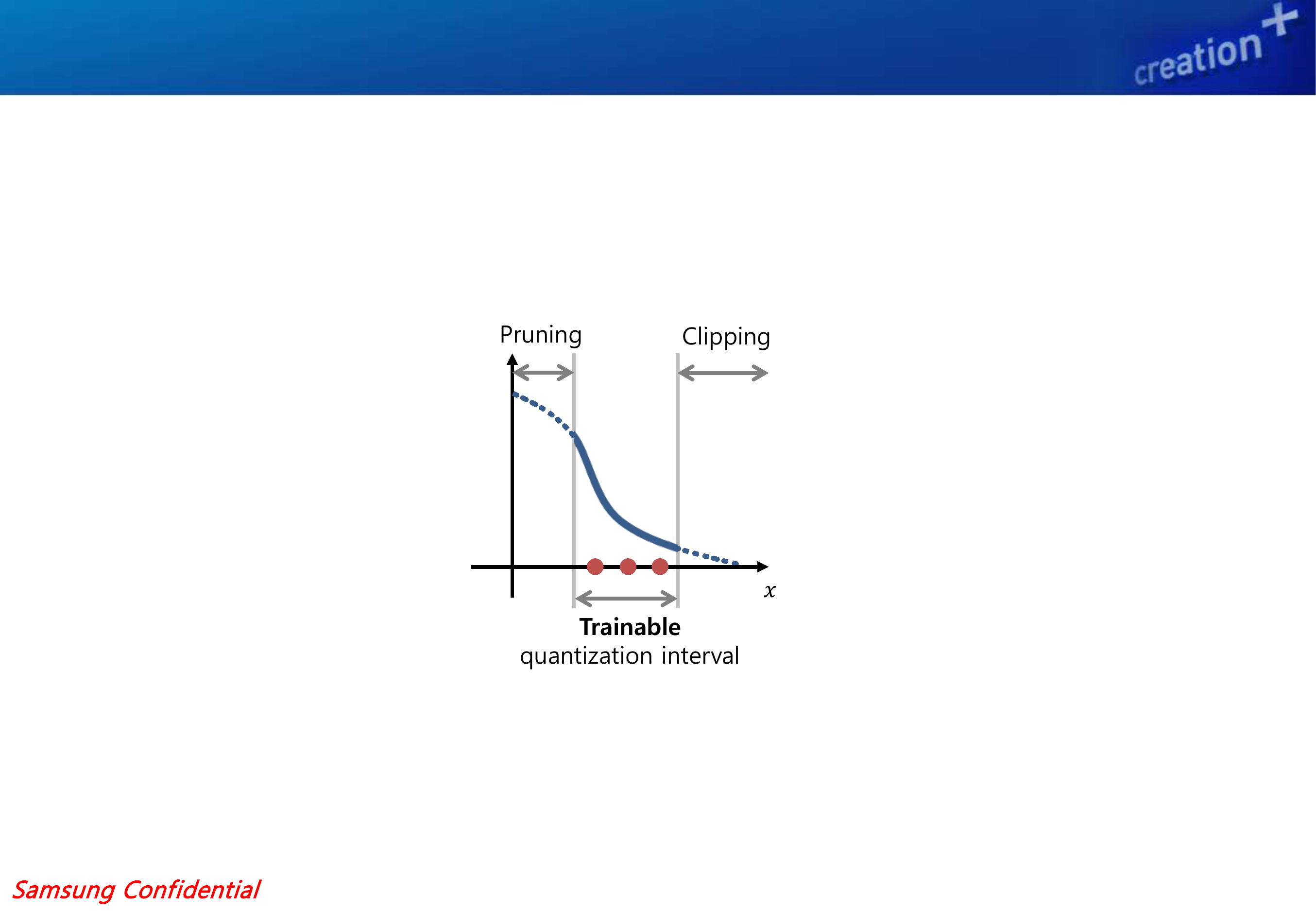} &
\includegraphics[width=0.56\linewidth]{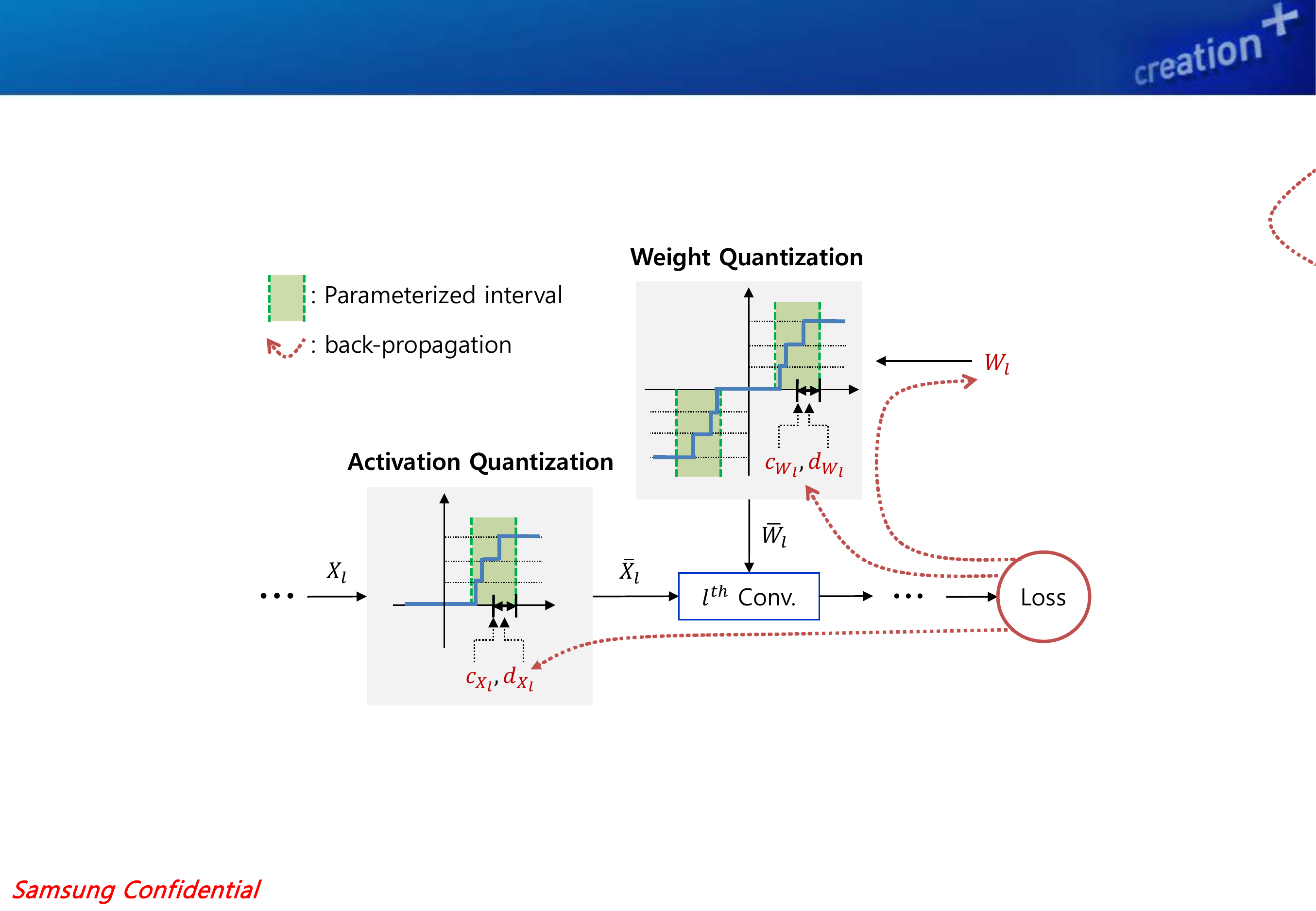} \\
(a) Quantization Interval & (b) A convolutional layer of our low bit-width network \\  
\end{tabular}  
  \caption{Illustration of our trainable quantizer. (a) Our trainable quantization interval, which performs pruning and clipping simultaneously. (b) The $l^{th}$ convolution layer of our low-precision network. Given bit-width, the quantized weights $\bar{W}_l$ and activations $\bar{X}_l$ are acquired using the parameterized intervals. The interval parameters ($c_{W_l}, d_{W_l}, c_{X_l}, d_{X_l}$) are trained jointly with the full-precision weights $W_l$ during backpropagation.}
  \label{fig:overview}
\end{figure*}

In order to maximally utilize the bit-wise operations for convolution, we should quantize both weights and activations. Binarized-Neural-Network(BNN) \cite{hubara2016binarized} binarizes the weights and activations to $\{-1,+1\}$ in the same way as BC, and uses these binarized values for computing gradients. XNOR-Net \cite{rastegari2016xnor} further conducts activation binarization with a scaling factor where the scaling factor is obtained in a closed form solution. DoReFa-Net \cite{zhou2016dorefa} performs a bit-wise operation for convolution by quantizing both weights and activations with multiple-bits rather than performing bipolar quantization. They adopt various activation functions to bound the activation values. The weights are transformed by the hyperbolic tangent function and then normalized with respect to the maximum values before quantization. Half-Wave-Gaussian-Quantization (HWGQ) \cite{cai2017deep} exploits the statistics of activations and proposes variants of ReLU activation which constrain the unbounded values.  Both DoReFa-Net and HWGQ use upper bounds for the activation but they are fixed during training and their quantizations are not learned as done in our model.

Several recent work \cite{zhang2018lq, wang2018two, zhuang2018towards, choi2018pact} proposed highly accurate low bit-width models by considering both weight and activation quantization. LQ-Nets \cite{zhang2018lq} allows floating-point values to represent the basis of $K$-bit quantized value instead of the standard basis $[1, 2, ..., 2^{K-1}]$, and learn the basis for weights and activations of each layer or channel by minimizing the quantization error. On the other hand, our trainable quantizer estimates the optimal quantization interval, which is learned in terms of minimizing the output task loss rather than minimizing the quantization error. Furthermore, the LQ-Nets has to use a few floating-point multiplication for computing convolution due to the floating-point basis, but our convolution can use shift operation instead of multiplication because all of our quantized values are integers. 

Wang et al.~\cite{wang2018two} proposes a two-step quantization (TSQ) that decomposes activation and weight quantization steps, which achieves good performance with AlexNet and VGGNet architectures. However, TSQ adopts layerwise optimization for the weight quantization step, and thus is not applicable to ResNet architecture which includes skip connections. Contrarily, our quantizer is applicable to any types of network architectures regardless of skip connection. Zhuang \etal \cite{zhuang2018towards} proposes a two-stage and progressive optimization method to obtain good initialization to avoid the network to get trapped in a poor local minima. We adopt their progressive strategy, which actually improves the accuracy especially for extremely low bit-width network, i.e., 2-bit. PACT \cite{choi2018pact} proposes a parameterized clipping activation function where the clipping parameter is obtained during training. However, it does not consider pruning and the weight quantization is fixed as in DoReFa-Net. A couple of recent quantization methods use Bayesian approches; Louizos et al. \cite{louizos2017bayesian} propose Bayesian compression, which determines optimal number of bit precision per layer via the variance of the estimated posterior, but does not cluster weights. Achterhold et al. \cite{achterhold2018variational} propose a variational inference framework to learn networks that quantize well, using multi-modal quantizing priors with peaks at quantization target values. Yet neither approaches prune activations or learn the quantizer itself as done in our work.  

We propose the trainable quantization interval which performs pruning and clipping simultaneously during training, and by applying this parameterization both for weight and activation quantization we keep the classification accuracy on par with the one of the full-precision network while significantly reducing the bit-width.


\section{Method}
\label{sec3}

The trainable quantization interval has quantization range within the interval, and prunes and clips ranges out of the interval. We apply the trainable quantization intervals to both activation and weight quantization and optimize them with respect to the task loss (Fig. \ref{fig:overview}). In this section, we first review and interpret the quantization process for low bit-width networks, and then present our trainable quantizer with quantization intervals.

\subsection{Quantization in low bit-width network}
\label{sec3.1}

\begin{figure*}[t]
  \centering
\begin{tabular}{ccc}
\includegraphics[width=0.25\linewidth]{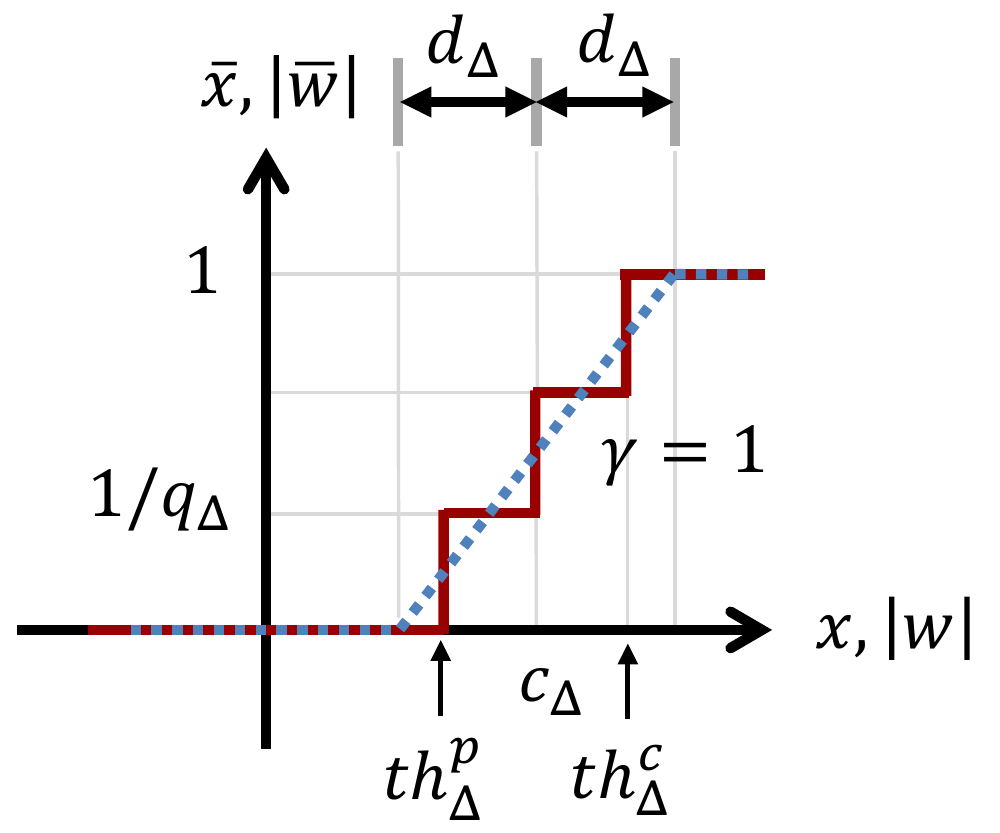} &
\includegraphics[width=0.25\linewidth]{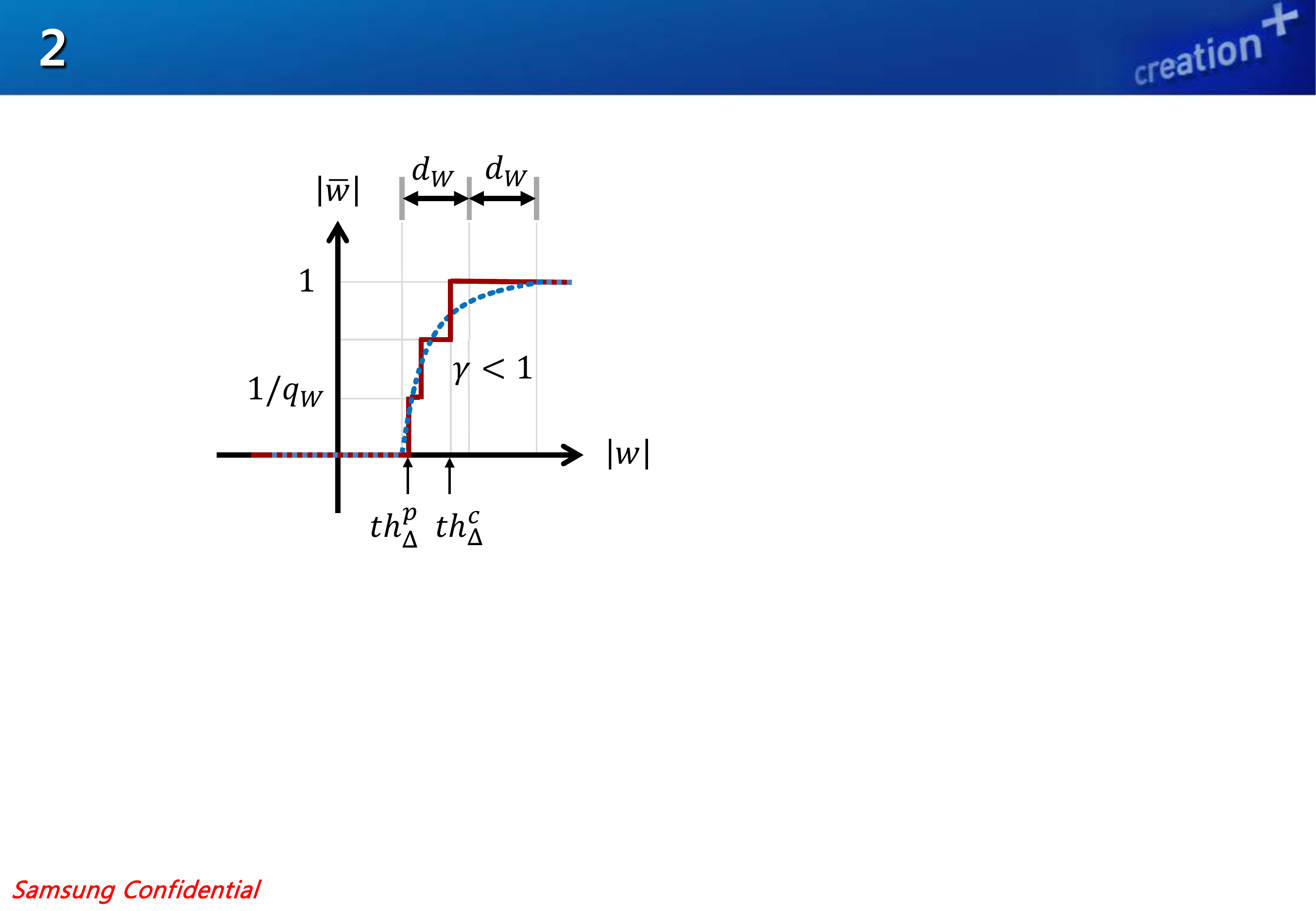} &
\includegraphics[width=0.25\linewidth]{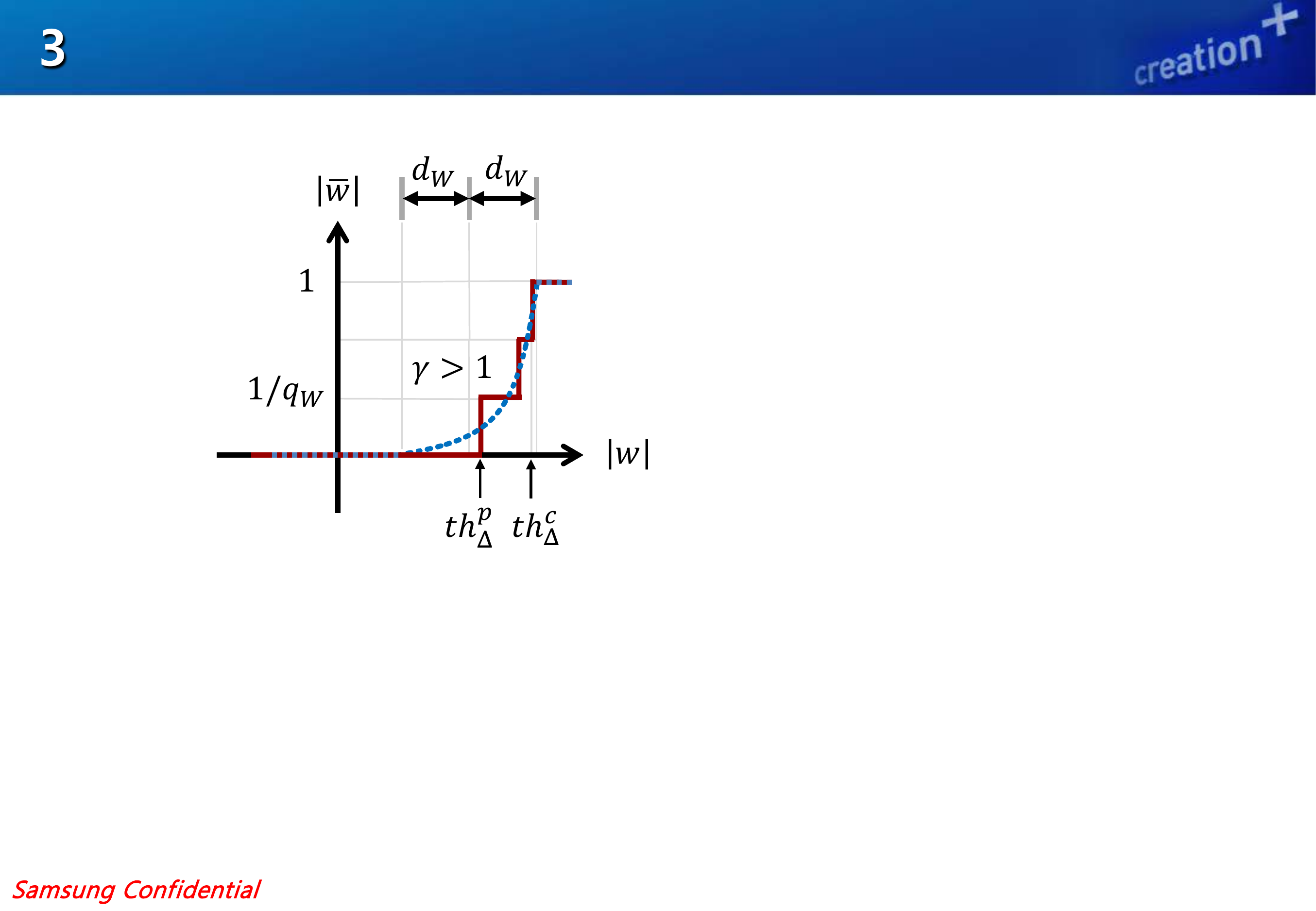}  \\
(a)   & (b)  & (c) \\  
\end{tabular}  
  \caption{A quantizer as a combination of a transformer and a discretizer with various $\gamma$ where (a) $\gamma=1$, (b) $\gamma<1$, and (c) $\gamma>1$. The blue dotted lines indicate the transformers, and the red solid lines are their corresponding quantizers. The $th^p_{\Delta}$ and the $th^c_{\Delta}$ represent the pruning and clipping thresholds, respectively.}
  \label{fig:quant}
\end{figure*}

For the $l$-th layer of a full-precision convolutional neural network (CNN), the weight $\bf{W}_l$ is convolved with the input activation $\bf{X}_l$ where $\bf{W}_l$ and $\bf{X}_l$ are real-valued tensors. We denote the elements of $\bf{W}_l$ and $\bf{X}_l$ by $w_l$ and $x_l$, respectively. For notational simplicity, we drop the subscript $l$ in the following.  Reducing bit-widths inherently involves a quantization process, where we obtain the quantized weight $\bar{w} \in \bar{\bf{W}}$ and the quantized input activation $\bar{x} \in \bar{\bf{X}}$ via quantizers,
\begin{equation}   
\begin{array}{rl}
Q_{W} : & w \myrightarrow{T_W} \hat{w} \myrightarrow{D}  \bar{w} \\
Q_{X}:   & x \myrightarrow{T_X} \hat{x} \myrightarrow{D} \bar{x}.  \\
\end{array} 
\label{eq:quantizer}
\end{equation}
A quantizer $Q_{\Delta}(\Delta \in \{\bf{W},\bf{X}\})$ is a composition of a transformer $T_{\Delta}$ and a discretizer $D$. The transformer maps the weight and activation values to $[-1,1]$ or $[0,1]$. The simplest example of the transformer is the normalization that divides the values by their maximum absolute values. Another example is a function of $\tanh(\cdot)$ for weights and clipping for activations \cite{zhou2016dorefa}. The discretizer maps a real-value $\hat{v}$ in the range $[-1,1]$ (or $[0,1]$) to some discrete value $\bar{v}$ as follows:
\begin{equation}   
\bar{v} = \frac{\lceil \hat{v} \cdot q_D \rfloor}{q_D}
\label{eq:discretizer}
\end{equation}
where $\lceil \cdot \rfloor$ is the round operation and $q_D$ is the discretization level.

In this paper, we parameterize the quantizer (transformer) to make it trainable rather than fixed. Thus the quantizers can be jointly optimized together with the neural network model weights. We can obtain the optimal $\bar{\bf{W}}$ and $\bar{\bf{X}}$ that directly minimize the task loss (i.e., classification loss) of the entire network (Fig. \ref{fig:overview} (b)) rather than simply approximating the full-precision weight/activation ($\bf{W} \approx \bar{\bf{W}}$, $\bf{X} \approx \bar{\bf{X}}$) \cite{li2016ternary, cai2017deep} or the convolutional outputs  ($\bf{W} \ast \bf{X}  \approx \bar{\bf{W}} \ast \bar{\bf{X}}$) \cite{rastegari2016xnor, wang2018two} where $\ast$ denotes convolution operation.

\subsection{Trainable quantization interval}
\label{sec3.2}

To design the quantizers, we consider two operations: clipping and pruning (Fig. \ref{fig:overview} (a)). The underlying idea of clipping is to limit the upper bound for quantization \cite{zhou2016dorefa, choi2018pact}. Decreasing the upper bound increases the quantization resolution within the bound so that the accuracy of the low bit-width network can increase. On the other hand, if the upper bound is set too low, accuracy may decrease because too many values will be clipped. Thus, setting a proper clipping threshold is crucial for maintaining the performance of the networks. Pruning removes low-valued weight parameters \cite{han2015deep}. Increasing pruning threshold helps to increase the quantization resolution and reduce the model complexity, while setting pruning threshold too high can cause the performance degradation due to the same reason as the clipping scheme does.

We define the quantization interval to consider both pruning and clipping. To estimate the interval automatically, the intervals are parameterized by $c_{\Delta}$ and $d_{\Delta}$ ($\Delta \in \{\bf{W}, \bf{X}\}$) for each layer where $c_{\Delta}$ and $d_{\Delta}$ indicate the center of the interval and the distance from the center, respectively. Note that this is simply a design choice and other types of parameterization, such as parameterization with lower and upper bound, are also possible.

Let us first consider the weight quantization. Because weights contain both positive and negative values, the quantizer has symmetry on both positive and negative sides. Given the interval parameters $c_W$ and $d_W$, we define the transformer $T_W$ as follows:

\begin{equation}   
\hat{w} = \left\{\begin{array}{cl}
0 & |w| < c_W-d_W \\
\textrm{sign}(w) & |w| > c_W+d_W\\
(\alpha_W |w| + \beta_W)^{\gamma} \cdot \textrm{sign}(w) & otherwise,
\end{array} \right.
\label{eq:weightquant}
\end{equation}
where $\alpha_W=0.5/d_W$, $\beta_W=-0.5c_W/d_W+0.5$ and the $\gamma$ is another trainable parameter of the transformer. That is, the quantizer is designed by the interval parameters $c_W$, $d_W$ and $\gamma$ which are trainable. The non-linear function with $\gamma$ considers the distribution inside the interval. The graphs in Fig. \ref{fig:quant} show the transformers (dotted blue lines) and their corresponding quantizers (solid red lines) with various $\gamma$. If $\gamma=1$, then the transformer is a piecewise linear function where the inside values of the interval are uniformly quantized (Fig. \ref{fig:quant} (a)). The inside values can be non-uniformly quantized by adjusting $\gamma$ (Fig. \ref{fig:quant} (b, c)). $\gamma$ could be either set to a fixed value or trained. We demonstrate the effects of $\gamma$ in the experiment. If $\gamma \ne 1$, the function is complex to be calculated. However, the weight quantizers are removed after training and we use only the quantized weights for inference. For this reason, this complex non-linear function does not decrease the inference speed at all. The actual pruning threshold $th^p_{\Delta}$ and clipping threshold  $th^c_{\Delta}$ vary according to the parameters $c_W$, $d_W$ and $\gamma$ as shown in Fig. \ref{fig:quant}. For example, the $th^p_{\Delta}$ and $th^c_{\Delta}$ in the case of $\gamma=1$ are derived as follows:
\begin{equation} \label{eq:th} 
\begin{array}{l}
th^p_{\Delta} = c_{\Delta}+d_{\Delta}+0.5d_{\Delta}/q_{\Delta} \\
th^c_{\Delta} = c_{\Delta}-d_{\Delta}-0.5d_{\Delta}/q_{\Delta}. \\
\end{array} 
\end{equation}
Note that the number of quantization levels $q_W$ for the weight can be computed as $q_W=2^{N_W-1}-1$ (one-side, except 0), given the bit-width $N_W$. Thus, the 2-bit weights are actually ternary $\{-1,0,1\}$. 

The activations fed into a convolutional layer are non-negative due to ReLU operation. For activation quantization, a value larger than $c_{X}+d_X$ is clipped and mapped to 1 and we prune a value smaller than $c_X-d_X$ to 0. The values in between are linearly mapped to $[0,1]$, which means that the values are uniformly quantized in the quantization interval. Unlike the weight quantization, activation quantizaiton should be conducted on-line during inference, thus we fix the $\gamma$ to 1 for fast computation. Then, the transformer $T_X$ for activation is defined as follows (Fig. \ref{fig:quant} (a)):

\begin{equation}   
\hat{x} = \left\{\begin{array}{cl}
0 & x < c_X-d_X \\
1 & x > c_X+d_X \\
\alpha_X x + \beta_X & otherwise,
\end{array} \right.
\label{eq:activquant}
\end{equation}
where $\alpha_X=0.5/d_X$ and $\beta_X=-0.5c_X/d_X+0.5$. Given the bit-width $N_X$ of activation, the number of quantization levels $q_X$ (except 0) can be computed as $q_X=2^{N_X}-1$; i.e., the quantized values are $\{0,1,2,3\}$ for 2-bit activations.

We use stochastic gradient descent for optimizing the parameters of both the weights and the quantizers. The transformers are piece-wise differentiable, and thus we can compute the gradient with respect to the interval parameters $c_{\Delta}$, $d_{\Delta}$ and $\gamma$. We use straight-through-estimator \cite{bengio2013estimating, zhou2016dorefa} for the gradient of the discretizers. 

Basically, our method trains the weight parameters jointly with quantizers. However, it is also possible to train the quantizers on a pre-trained network with full-precision weights. Surprisingly, training only the quantizer without updating weights also yields reasonably good accuracy, although its accuracy is lower than that of joint training (See Fig. \ref{fig:cifar-100}). 

We describe the pseudo-code for training and deploying our low bit-width model in Algorithm \ref{algo}.

\begin{algorithm}[t]
\caption{Training low bit-width network using parameterized quantizers} \label{algo}
\renewcommand{\algorithmicrequire}{\textbf{Input:}}
\renewcommand{\algorithmicensure}{\textbf{Output:}}
\begin{algorithmic} [1]
\Require Training data
\Ensure A low bit-width model with quantized weights $\{\bar{w}_{l}\}_{l=1}^{L}$ and activation quantizers $\{c_{X_l}, d_{X_l}\}_{l=1}^{L}$ 
\Procedure{Training}{} 
\State Initialize the parameter set $\{{P_l}\}_{l=1}^{L}$ where $P_l=\{{w}_{l}, c_{W_l}, d_{W_l}, \gamma_l, c_{X_l}, d_{X_l}\}$
\For {$l=1, ..., L$}
\State Compute ${\bar{w}_l}$ from ${w_l}$ using Eq. \ref{eq:weightquant} and Eq. \ref{eq:discretizer}
\State Compute ${\bar{x}_l}$ from ${x_l}$ using Eq. \ref{eq:activquant} and Eq. \ref{eq:discretizer}
\State Compute ${\bar{w}_l} \ast {\bar{x}_l}$
\EndFor
Compute the loss $\ell$
\State Compute the gradient w.r.t. the output $\partial \ell / \partial x_{L+1}$
\For {$l=L, ..., 1$}
\State Given $\partial \ell / \partial{x}_{l+1}$, 
\State Compute the gradient of the parameters in $P_l$ 
\State Update the parameters in $P_l$
\State Compute $\partial \ell / \partial{x}_{l}$
\EndFor
\EndProcedure
\Procedure{Deployment}{} 
\For {$l=1, ..., L$}
\State Compute ${\bar{w}_l}$ from ${w_l}$ using Eq. \ref{eq:weightquant} and Eq. \ref{eq:discretizer}
\EndFor
\State Deploy the low bit-width model $\{{w}_{l}, c_{X_l}, d_{X_l}\}_{l=1}^{L}$
\EndProcedure
\end{algorithmic}
\end{algorithm}

\begin{table*}[ht] 
  \caption{Top-1 accuracy (\%) on ImageNet. Comparion with the existing methods on ResNet-18, -34 and AlexNet. The `FP' represents the full-precision (32/32-bit) accuracy in our implementation.}
  \label{Table:ImageNet}
  \centering
  \begin{tabular}{l|cccc|cccc|cccc}
  \toprule
  \multirow{3}{*}{Method}   &\multicolumn{4}{c|}{ResNet-18 (FP: \textbf{70.2})}  & \multicolumn{4}{c|}{ResNet-34 (FP: \textbf{73.7})}  &\multicolumn{4}{c}{AlexNet (FP: \textbf{61.8})}\\
   \cmidrule{2-13}      
         & \multicolumn{12}{c}{Bit-width (A/W) } \\
   &5/5 & 4/4 & 3/3 & 2/2&  5/5 & 4/4 & 3/3 & 2/2 & 5/5 & 4/4 & 3/3 & 2/2\\  
  \midrule   
  \midrule  
  \textbf{QIL (Ours)}\footnotemark[2]  &  \textbf{70.4} & \textbf{70.1} & \textbf{69.2} &  \textbf{65.7} &\textbf{73.7} & \textbf{73.7} &           
                           \textbf{73.1} & \textbf{70.6} & \textbf{61.9} &\textbf{62.0} & \textbf{61.3} & \textbf{58.1}\\
  LQ-Nets \cite{zhang2018lq}           & - & 69.3 & 68.2 & 64.9 & - & - & 71.9 & 69.8 & - & - & - & 57.4 \\  
  PACT \cite{choi2018pact}              & 69.8 & 69.2 & 68.1 & 64.4 & - & - & - & - & 55.7 & 55.7 & 55.6 & 55.0\\
  DoReFa-Net \cite{zhou2016dorefa}  & 68.4 & 68.1 & 67.5 & 62.6 & - & -& -& -& 54.9 & 54.9 & 55.0 &  53.6\\
  ABC-Net \cite{lin2017towards}        & 65.0 & - & 61.0 & - & 68.4 & - & 66.7 & - & - & - & - & - \\
  BalancedQ \cite{zhou2017balanced} & - & - & - & 59.4 & - & - & - & - & - & - & - & 55.7 \\
  TSQ\footnotemark[2] \cite{wang2018two} & - & - & - & - & - & - & - & - & - & - & - & 58.0\\
  SYQ\footnotemark[2] \cite{faraone2018syq} & - & - & - & - & - & - & - & - & - & - & - &55.8\\
   Zhuang \etal\cite{zhuang2018towards}   & - & - & - & - & - & - & - & - & - & 58.1& - & 52.5\\
   WEQ \cite{park2017weighted} & - & - & - & - & - & - & - & - & - & 55.9 & 54.9 & 50.6\\
  \bottomrule                                                                    
  \end{tabular}
\end{table*}

\section{Experiment results}
\label{sec4}

To demonstrate the effectiveness of our trainable quantizer, we evaluated it on the ImageNet \cite{russakovsky2015imagenet} and the CIFAR-100 \cite{krizhevsky2009learning} datasets.

\subsection{ImageNet}

The ImageNet classification dataset \cite{russakovsky2015imagenet} consists of 1.2M training images from 1000 general object classes and 50,000 validation images. We used various network architectures such as ResNet-18, -34 and AlexNet for evaluation. 

\paragraph{Implementation details }

We implement our method using PyTorch with multiple GPUs. We use original ResNet architecture \cite{he2016identity} without any structural change for ResNet-18 and -34. For AlexNet \cite{krizhevsky2012imagenet}, we use batch-normalization layer after each convolutional layer and remove the dropout layers and the LRN layers while preserving the other factors such as the number and the sizes of filters. In all the experiments, the training images are randomly cropped and resized to $224 \times 224$, and horizontally flipped at random. We use no further data augmentation. For testing, we use the single center-crop of $224 \times 224$. We used stochastic gradient descent with the batch size of 1024 (8 GPUs), the momentum of 0.9, and the weight decay of 0.0001 for ResNet (0.0005 for AlexNet). We trained each full-precision network up to 120 epochs where the learning rate was initially set to 0.4 for ResNet-18 and -34 (0.04 for AlexNet), and is decayed by a factor of 10 at 30, 60, 85, 95, 105 epochs. We finetune low bit-width networks for up to 90 epochs where the learning rate is set to 0.04 for ResNet-18 and -34 (0.004 for AlexNet) and is divided by 10 at 20, 40, 60, 80 epochs. We set the learning rates of interval parameters to be 100$\times$ smaller than those of weight parameters. We did not quantize the first and the last layers as was done in \cite{hubara2016quantized, zhou2016dorefa}.

\paragraph{Comparison with existing methods}

We evaluate our learnable quantization method with existing methods, by quoting the reported top-1 accuracies from the original papers (Table \ref{Table:ImageNet}). Our 5/5 and 4/4-bit models preserve the accuracy of full-precision model for all three network architectures (ResNet-18, -34 and AlexNet). For 3/3-bit, the accuracy drops only by 1\% for ResNet-18 and by 0.6\% for ResNet-34. If we further reduced the bit-width to 2/2-bit, the accuracy drops by 4.5\% (ResNet-18) and 3.1\% compared to the full-precision. Compared to the second best method (LQ-Nets \cite{zhang2018lq}), our 3/3 and 2/2-bit models are around 1\% more accurate for ResNet architectures. 
For AlexNet, our 3/3-bit model drops the top-1 accuracy only by 0.5\% with respect to full-precision which beats the second best by large margin of 5.7\%. For 2/2-bit, accuracy drops by 3.7\% which is almost same accuracy with TSQ \cite{wang2018two}. Note that TSQ used layerwise optimization, which makes it difficult to apply to the ResNet architecture with skip connections. However, our method is general and is applicable any types of network architecture.

\paragraph{Initialization}

\begin{table} [t]
  \caption{The top-1 accuracy (\%) of low bit-width networks on ResNet-18 with direct and progressive finetuning. The 5/5-bit network was finetuned only from full-precision network.}
  \label{Table: initialization}
  \centering
  \begin{tabular}{cccccc}
   \toprule
   \multirow{2}{*}{Initialization}  & \multicolumn{5}{c}{Bit-width (A/W)}\\  
   \cmidrule{2-6} & 32/32 & 5/5 & 4/4 & 3/3  & 2/2   \\
    \midrule   
    Direct & \multirow{2}{*}{70.2} & 70.4 & 69.9 & 68.7 & 56.0  \\
    Progressive & & - & 70.1 & 69.2 & 65.7   \\
   \bottomrule                                                                    
  \end{tabular}
\end{table}

\begin{table} [t]
  \caption{Joint training vs. Quantizer only. The top-1 accuracy (\%) with ResNet-18.}
  \label{Table:jointvsquantonly}
  \centering
  \begin{tabular}{cccccc}
   \toprule
   \multirow{2}{*}{Initialization}  & \multicolumn{5}{c}{Bit-width (A/W)}\\  
   \cmidrule{2-6} & 32/32 & 5/5 & 4/4 & 3/3  & 2/2   \\
    \midrule   
    Joint training & \multirow{2}{*}{70.2} & 70.4 & 70.1 & 69.2 & 65.7   \\
    Quantizer only &  & 69.4 & 68.0 & 62.0 & 20.9  \\
   \bottomrule                                                                    
  \end{tabular}
\end{table}

Good initialization of parameters improves the accuracy of trained low bit-width networks \cite{lin2017towards, mckinstry2018discovering}.  We adopt finetuning approach for good initialization. In \cite{lin2017towards}, they progressively conduct the quantization from higher bit-widths to lower bit-widths for better initialization. We compare the results of direct finetuning of full-precision network with progressively finetuning from 1-bit higher bit-width network (Table \ref{Table: initialization}). For progressive finetuning, we sequentially train the 4/4, 3/3, and 2/2-bit networks (i.e, FP $\rightarrow$ 5/5 $\rightarrow$ 4/4 $\rightarrow$ 3/3 $\rightarrow$ 2/2 for 2/2-bit network). Generally, the accuracies of progressive finetuning are higher than those of direct finetuning. The progressive finetuning is crucial for 2/2-bit network (9.7\% point accuracy improvement), but has marginal impact on 4/4 and 3/3-bit networks (only 0.2\% and 0.5\% point improvements, respectively).

\paragraph{Joint training vs. Quantizer only}

The weight parameters can be optimized jointly with quantization parameters or we can optimize only the quantizers while keeping the weight parameters fixed. Table \ref{Table:jointvsquantonly} shows the top-1 accuracy with ResNet-18 network on the both cases. Both the cases utilize the progressive finetuning. The joint training of quantizer and weights works better than training the quantizer only, which is consistent with our intuition. The joint training shows graceful performance degradation while reducing bit-width, compared to training of only the quantizers. Nevertheless, the accuracy of the quantizers is quite promising with 4-bit or higher bit-width. For example, the accuracy drop with 4/4-bit model is only 2.1\% (70.1\%$\rightarrow$68.0\%).


\footnotetext{\footnotemark[2] With this mark in Table \ref{Table:ImageNet} and \ref{Table:ImageNetResNet-18-WeiQuant}, the 2-bit of weights is ternary $\{-1,0,1\}$, otherwise it is 4-level.}

\paragraph{Pruning ratio}

\begin{figure} [t]
\centering
\includegraphics[width=0.85\linewidth]{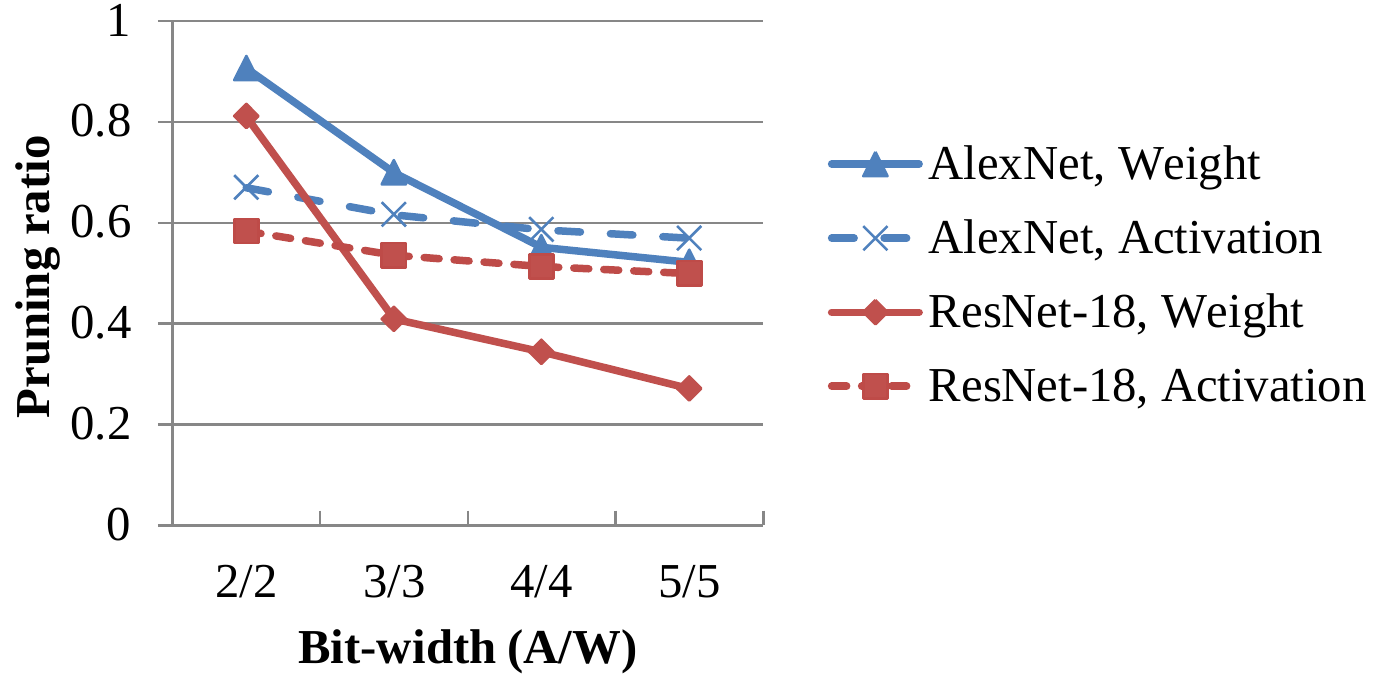} 
\caption{Average pruning ratio of weights and activations on AlexNet and ResNet-18 with various bit-widths}
\label{fig:zero-avg}
\end{figure}

\begin{figure*} [t] 
\centering
\begin{tabular}{cc}
\includegraphics[width=0.42\linewidth]{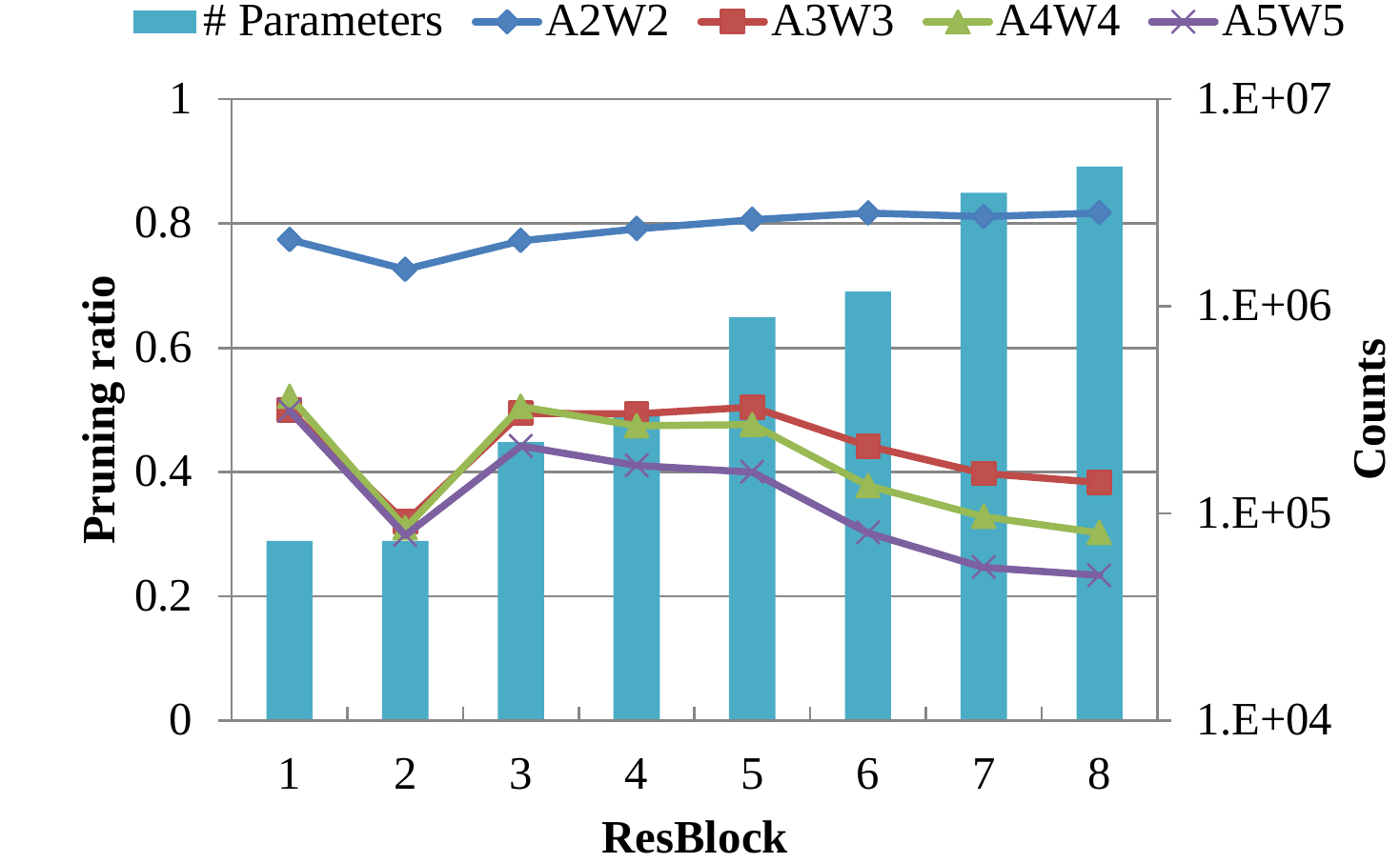} &
\includegraphics[width=0.42\linewidth]{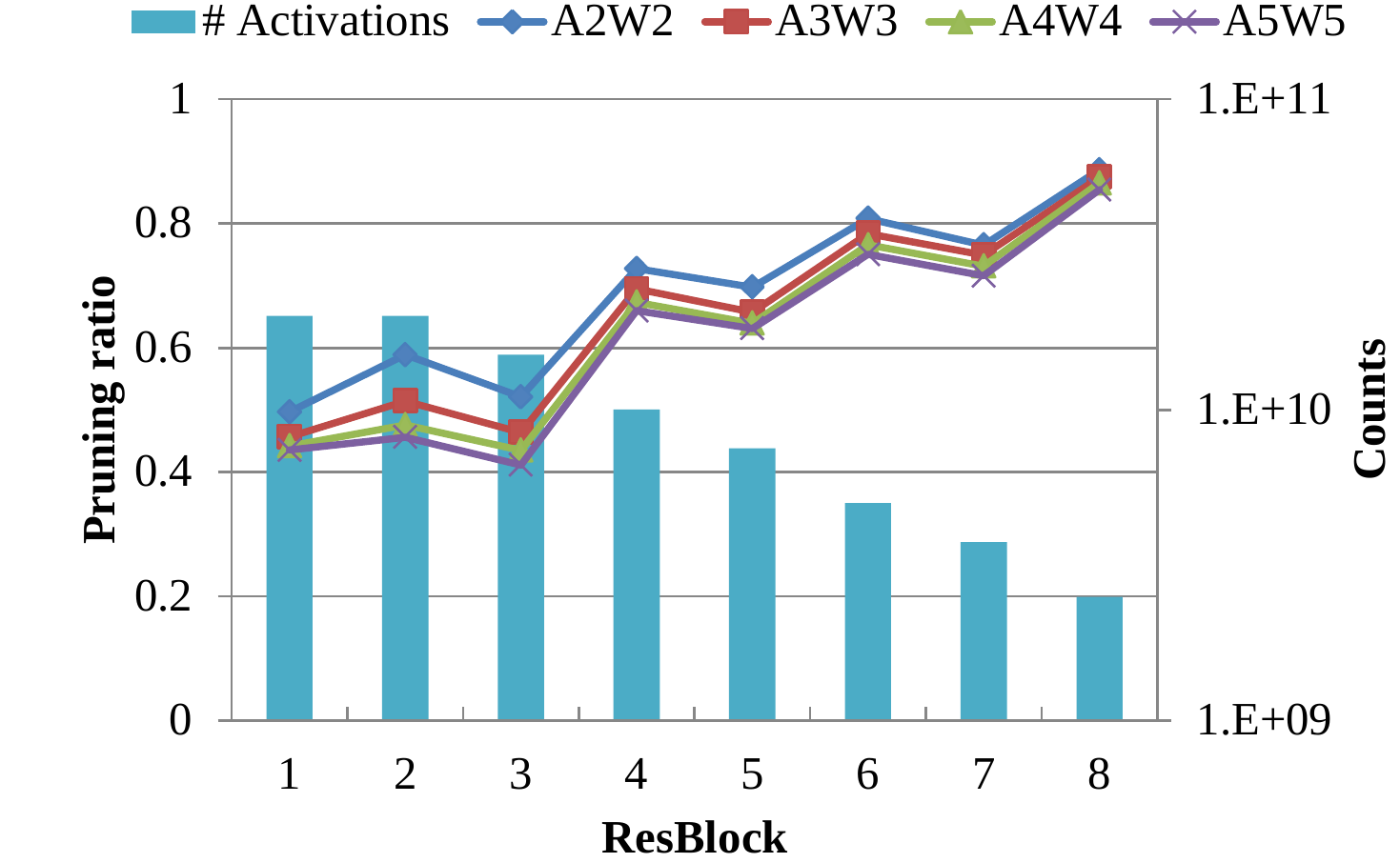} \\
(a) Weight  & (b)  Activation \\  
\end{tabular}
\caption{Blockwise pruning ratio of (a) weight and (b) activation for each ResBlock with ResNet-18. The bar graph shows the number of weights or activations for each ResBlock on a log scale.  }
\label{fig:zero-layerwise}
\end{figure*}

\begin{figure*} [t] 
\centering
\begin{tabular}{cccc}
\includegraphics[height=0.21\linewidth]{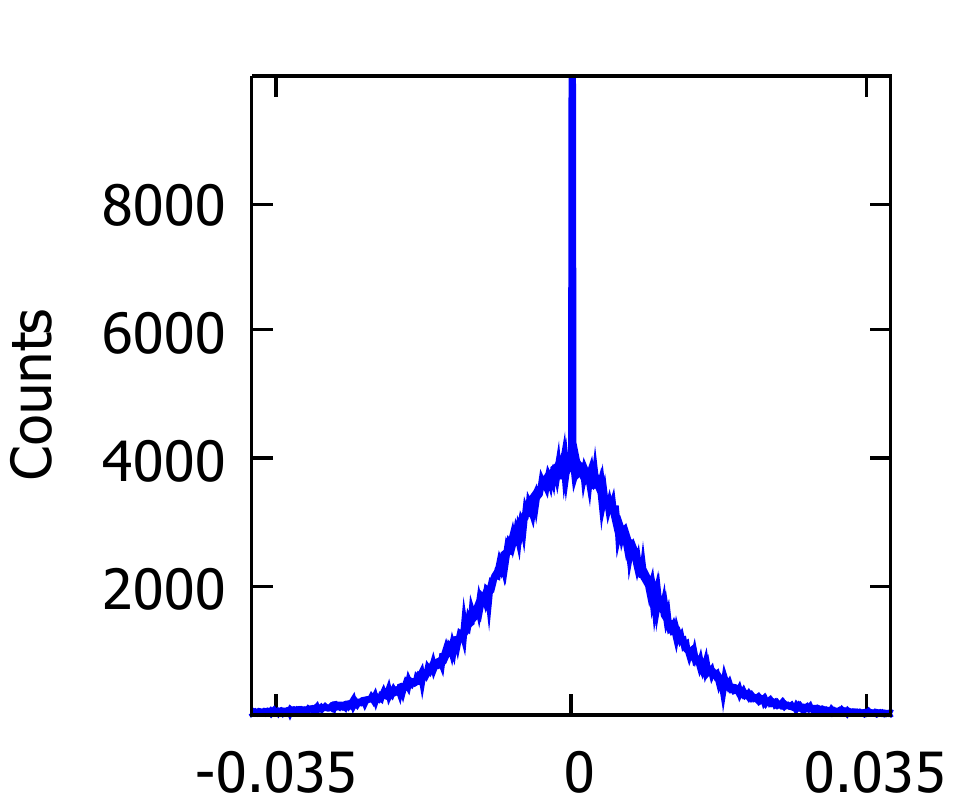} &
\includegraphics[height=0.21\linewidth]{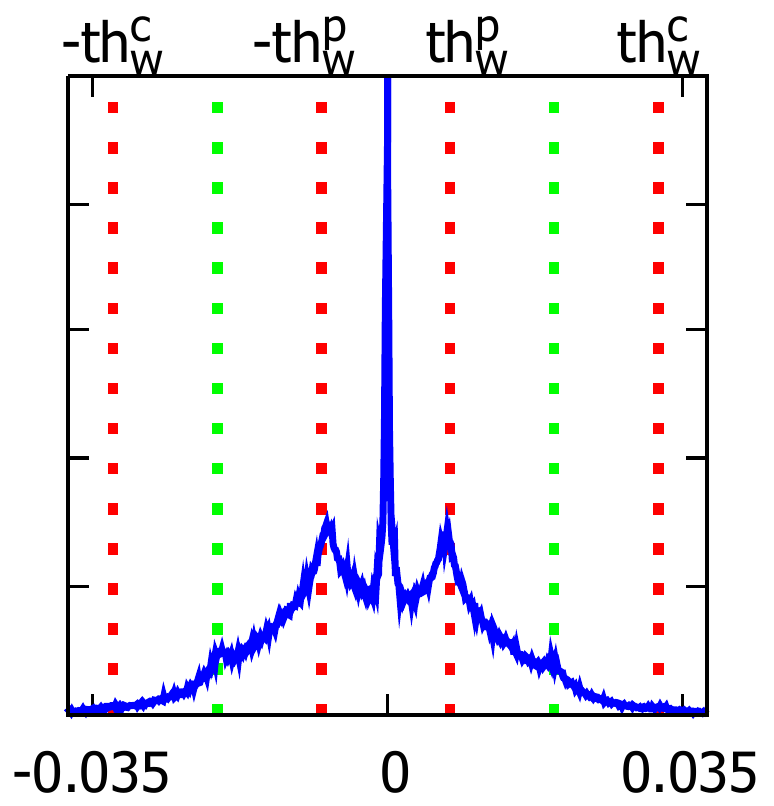} &
\includegraphics[height=0.21\linewidth]{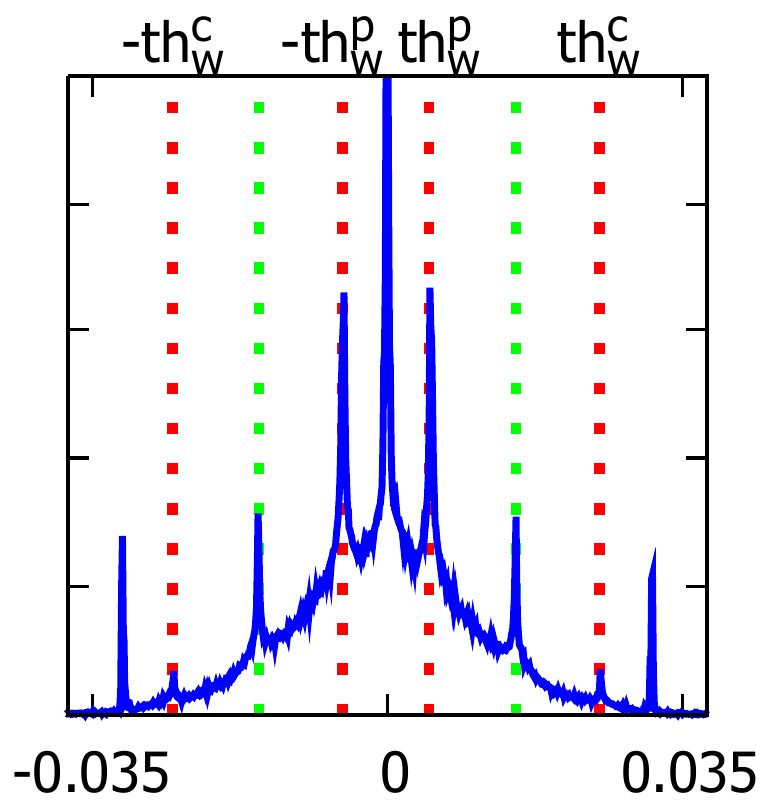} &
\includegraphics[height=0.21\linewidth]{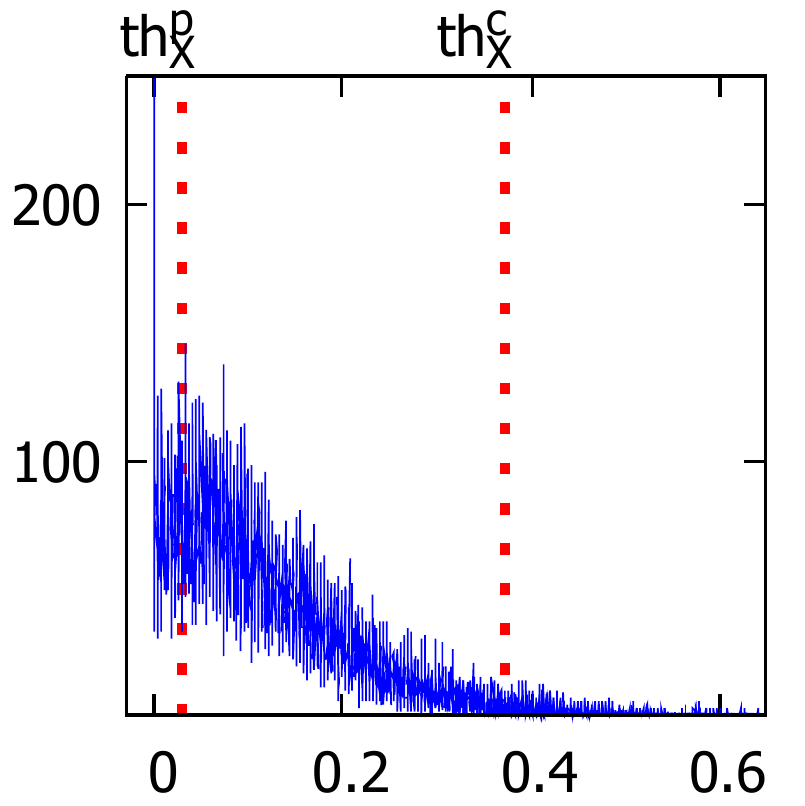} \\
\hspace{4ex}(a) weight, initial & (b) weight, epoch 11 & (c) weight, epoch 90& \hspace{3ex}(d) activation, epoch 90\\ 
\end{tabular}
\caption{Weight and activation distributions at the 3rd layer of the 3/3-bit AlexNet. The 3/3-bit network is finetuned from the pretrained full-precision network. The figures show the distributions of (a) initial weights, (b) weights at epoch 11, (c) weight at epoch 90 and (d) activation at epoch 90. The $th^p_{\Delta}$ and the $th^c_{\Delta}$ represent the pruning and the clipping thresholds, respectively. }
\label{fig:dist}
\end{figure*}

In order to see the pruning effect of our quantizer, we compute the pruning ratio which is the number of zeros over the total number of weights or activations. Fig. \ref{fig:zero-avg} shows the average pruning ratios of the entire network for ResNet-18 and AlexNet. As expected, the pruning ratio increases as the bit-width decreases. If the bit-width is high, the quantization interval can be relaxed due to the high quantization resolution. However, as the bit-width decreases, more compact interval should be found to maintain the accuracy, and as a result, the pruning ratio increases. For 2/2-bit network, 91\% and 81\% of weights are pruned on average for AlexNet and ResNet-18, respectively (Fig. \ref{fig:zero-avg}). The AlexNet is more pruned than the ResNet-18 because the AlexNet has fully-connected layers which have 18$\sim$64 times larger parameters than the convolutional layers and the fully-connected layers are more likely pruned. The activations are less affected by the bit-width. Fig. \ref{fig:zero-layerwise} shows the blockwise pruning ratio of ResNet-18. We compute the pruning ratio for each ResBlock which consists of two convolutional layers and a skip connection. For activations, the upper layers are more likely to be pruned, which may be because more abstraction occurs at higher layers.

\paragraph{Training $\gamma$}

For weight quantization, we designed the transformer with $\gamma$ (Eq. \ref{eq:weightquant}) which considers the distribution inside the interval. We investigated the effects of training $\gamma$. Table \ref{Table: gamma} shows the top-1 accuracies according to the various $\gamma$ for 3/3 and 2/2-bit AlexNet. We report both the trainable $\gamma$ and fixed $\gamma$. For 3/3-bit model, the trainable $\gamma$ does not affect the model accuracy; i.e., 61.4\% with trainable $\gamma$ and 61.3\% with $\gamma=1$. However, the trainable $\gamma$ is effective for 2/2-bit model which improves the top-1 accuracy by 0.9\% compared with $\gamma=1$ while the accuracies of the fixed $\gamma$ of 0.5 and 0.75 are similar with $\gamma=1$. The fixed $\gamma$ of 1.5 degrades the performance. We also evaluated trainable $\gamma$ with various network models for 2/2-bit (Table \ref{Table: gamma-arch}). The trainable $\gamma$ is less effective for ResNet-18 and -34.


\begin{table} [t]
  \caption{The top-1 accuray with various $\gamma$ on AlexNet}
  \label{Table: gamma}
  \centering
  \begin{tabular}{cccccc}
   \toprule
   \multirow{2}{*}{\shortstack[c]{Bit-width \\ (A/W) }}&  \multicolumn{4}{c}{Fixed $\gamma$} & \multirow{2}{*}{\shortstack[c]{Trainable \\ $\gamma$ }}\\
   \cmidrule{2-5} & 0.5 & 0.75  & 1.0 & 1.5  &  \\
    \midrule   
  3/3    & 61.0 & 61.3 & 61.3 & 61.0 & 61.4 \\    
  2/2    & 57.3 & 57.2 & 57.2 & 56.8 & 58.1  \\
   \bottomrule                                                                    
  \end{tabular}
\end{table}

\begin{table} [t]
  \caption{Trainable $\gamma$ with various network architecture for 2/2-bit}
  \label{Table: gamma-arch}
  \centering
  \begin{tabular}{cccccc}
   \toprule
   & AlexNet & ResNet-18 & ResNet-34 \\
   \midrule
Fixed $\gamma=1$ & 57.2 &  65.7 & 70.6 \\
Trainable $\gamma$ & 58.1 (+0.9) & 66.1 (+0.4) & 70.6 (+0.0)  \\
   \bottomrule                                                                    
  \end{tabular}
\end{table}

\paragraph{Weight quantization}

To demonstrate the effectiveness of our method on weight quantization, we only quantize the weights and leave the activations as full-precision. We measure the accuracy on ResNet-18 with various bit-widths for weights and compared the other existing methods (Table \ref{Table:ImageNetResNet-18-WeiQuant}). The accuracies of the networks quantized with our method are slightly higher than those obtained using the second-best method (LQ-Nets \cite{zhang2018lq}).

\paragraph{Distributions of weights and activations}

Fig. \ref{fig:dist} shows the distributions of weights and activations with different epochs. Since we train both weights and quantizers, the distributions of weights and quantization intervals are changing during training. Note that the peaks of the weight distribution appear at the transition of each quantized values. If the objective is to minimize the quantization errors, this distribution is not desirable. Since our objective is to minimize the task loss, during training the loss become different only when the quantized level of weights are different. Therefore the weight values move toward the transition boundaries. We also plot the activation distribution with the pruning and the clipping thresholds. (Fig. \ref{fig:dist} (d)).

\begin{table} [t] 
  \caption{Accuracy with ResNet-18 on ImageNet. The weights are quantized to low bits and the activations remain at 32 bits. The TTQ-B and TWN-B used the ResNet-18B \cite{li2016ternary} where the number of filters in each block is $1.5\times$ of the ResNet-18. }
  \label{Table:ImageNetResNet-18-WeiQuant}
  \centering
  \begin{tabular}{clccc}
   \toprule
   \multirow{2}{*}{\shortstack[c]{Bit-width \\ (A/W) }}  & \multirow{2}{*}{Method} & \multicolumn{2}{c}{Accuracy (\%)}    \\  
   & & Top-1 & Top-5 \\
    \midrule
    \midrule
    32/32                        & FP-ours                                                 & 70.2 & 89.6 \\
    \midrule
    \multirow{2}{*}{32/4} &  \textbf{QIL (Ours)}                                        & \textbf{70.3} & \textbf{89.5} \\
                                    &  LQ-Nets \cite{zhang2018lq}                      & 70.0 & 89.1\\  
    \midrule                                    
    \multirow{2}{*}{32/3} &  \textbf{QIL (Ours)}                                        & \textbf{69.9} & \textbf{89.3} \\
                                    & LQ-Nets \cite{zhang2018lq}                     & 69.3 & 88.8 \\
    \midrule 
    \multirow{5}{*}{32/2} &  \textbf{QIL\footnotemark[2](Ours)}                & \textbf{68.1} & \textbf{88.3} \\   
                                    & LQ-Nets \cite{zhang2018lq}                     & 68.0 & 88.0 \\
                                    & TTQ-B\footnotemark[2]    \cite{zhu2016trained}                & 66.6 & 87.2 \\         
                                    & TWN-B\footnotemark[2]  \cite{li2016ternary}                     & 65.3 & 86.2 \\                                                                                                                                                  
                                    & TWN\footnotemark[2]   \cite{li2016ternary}                       & 61.8 & 84.2 \\    
   \bottomrule                                                                    
  \end{tabular}
\end{table}

\subsection{CIFAR-100}

\begin{figure} [t]
\centering
\includegraphics[width=0.9\linewidth]{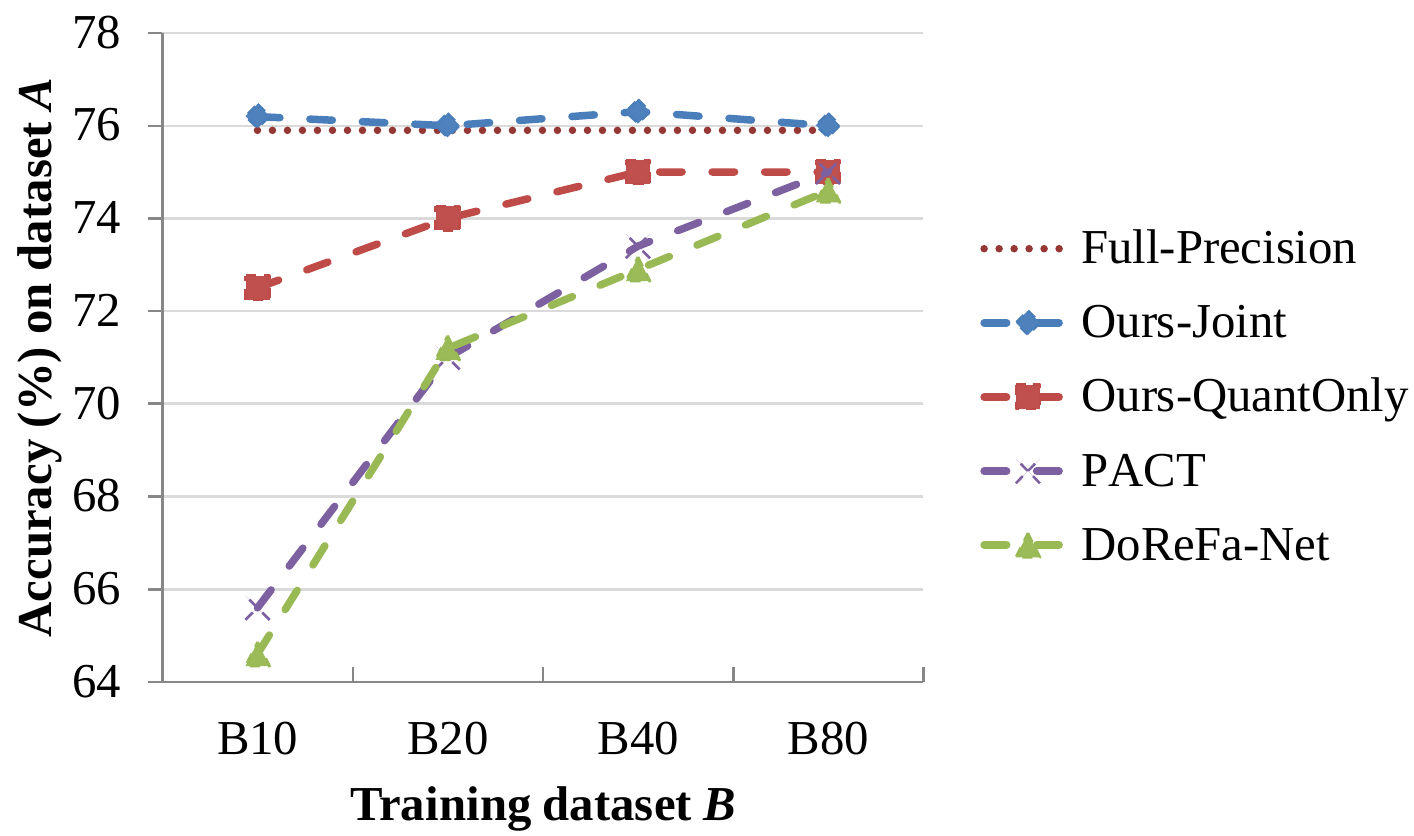} 
\caption{The accuracy on CIFAR-100 for low bit-width model training with heterogeneous dataset.}
\label{fig:cifar-100}
\end{figure}

In this experiment, we train a low bit-width network from a pre-trained full-precision network without the original training dataset. The motivation of this experiment is to validate whether it is possible to train the low bit-width model when the original training dataset is not given. To demonstrate this scenario, we used CIFAR-100 \cite{krizhevsky2009learning}, which consists of 100 classes containing 500 training and 100 validation images for each class where 100 classes are grouped into 20 superclasses (4 classes for each superclass). We divide the dataset into two disjoint groups $A$ (4 superclasses, 20 classes) and $B$ (16 superclasses, 80 classes) where $A$ is the original training set and $B$ is used as a heterogeneous dataset for training low bit-width model (4/4-bit model in this experiment). The $B$ is further divided into four subsets $B10$, $B20$, $B40$ and $B80$ ($B10 \subset B20 \subset B40 \subset B80 = B$) with 10, 20, 40 and 80 classes, respectively. First, we train the full-precision networks with $A$, then finetune the low bit-width networks with $B$ by minimizing the mean-squared-errors between the outputs of the full-precision model and the low bit-width model. We report the testing accuracy of $A$ for evaluation (Fig. \ref{fig:cifar-100}). We compare our method with other existing methods such as DoReFa-Net \cite{zhou2016dorefa} and PACT \cite{choi2018pact}. We carefully re-implemented these methods with pyTorch. Our joint training of weights and quantizers preserve the full-precision accuracy with all types of $B$. If we train the quantizer only with fixed weights, then the accuracies are lower than those of the joint training. Our method achieves better accuracy compared to DoReFa-Net and PACT. As expected, as the size of dataset increases ($B10 \rightarrow B20 \rightarrow B40 \rightarrow B80$), the accuracy improves. These are impressive results, since they show that we can quantize any given low bit-width networks can without access to the original data they are trained on.

\section{Conclusion}
\label{sec5}

We proposed a novel trainable quantizer with parameterized quantization intervals for training low bit-width networks. Our trainable quantizer performs simultaneous pruning and clipping for both weights and activations, while maintaining the accuracy of the full-precision network by learning appropriate quantization intervals. Instead of minimizing the quantization error with respect to the weights/activations of the full-precision networks as done in previous work, we train the quantization parameters jointly with the weights by directly minimizing the task loss. As a result, we achieved very promising results on the large scale ImageNet classification dataset. The 4-bit networks obtained using our method preserve the accuracies of the full-precision networks with various architectures, 3-bit networks yield comparable accuracy to the full-precision networks, and the 2-bit networks suffers from minimal accuracy loss. Our quantizer also achieves good quantization performance that outperforms the existing methods even when trained on a heterogeneous dataset, which makes it highly practical in situations where we have pretrained networks without access to the original training data. Future work may include more accurate parameterization of the quantization intervals with piecewise linear functions and use of Bayesian approaches.

{\small
\bibliographystyle{ieee}
\bibliography{egbib}
}

\end{document}